%% file: paper.tex
\documentclass[sigplan,11pt]{acmart}

\startPage{1}

\setcopyright{none}

\usepackage{booktabs}   
\usepackage{subcaption} 

\begin{document}

\title[]{DeepThin: A Self-Compressing Library for Deep Neural Networks}


\author{Matthew Sotoudeh}
\authornote{Work done while at Intel.}          
\affiliation{
  \institution{Intel/UC Davis}            
}
\email{masotoudeh@ucdavis.edu}          

\author{Sara S. Baghsorkhi}
\affiliation{
  \institution{Intel}           
}
\email{sara.s.baghsorkhi@intel.com}         

\begin{abstract}
  As the industry deploys increasingly large and complex neural networks to mobile devices, 
  more pressure is put on
  the memory and compute resources of those devices.  Deep compression, or compression of deep neural network weight
  matrices, is a technique to stretch resources for such scenarios. Existing compression methods cannot
  effectively compress models smaller than 1-2\% of their original size.
  We develop a new compression technique, DeepThin, building on existing research in the area of low rank factorization. We identify and break 
  artificial constraints imposed by low rank approximations by combining rank factorization with a reshaping process that adds nonlinearity to
  the approximation function.
  We deploy DeepThin as a plug-gable library integrated with TensorFlow that enables users to seamlessly compress models at different granularities. 
  We evaluate DeepThin on two state-of-the-art acoustic models, TFKaldi and DeepSpeech, comparing it to previous
  compression work (Pruning, HashNet, and Rank Factorization), empirical limit study approaches, and hand-tuned models.
  For TFKaldi, our DeepThin networks show better word error rates (WER) than competing methods at practically all tested compression rates,
  achieving an average of 60\% relative improvement over rank factorization, 57\% over pruning,
  23\% over hand-tuned same-size networks,  and 6\% over the computationally expensive
  HashedNets. For DeepSpeech, DeepThin-compressed networks achieve better test loss than all
  other compression methods, reaching a 28\% better result than rank factorization, 27\%
  better than pruning, 20\% better than hand-tuned same-size networks, and 12\% better than
  HashedNets.

  DeepThin also provide inference
  performance benefits in two ways: (1) by shrinking the application working sets, allowing the model to fit in a level of the cache/memory hierarchy
  where the original network was too large, and (2) by exploiting unique features of the technique to reuse many intermediate computations, reducing the total compute operations necessary.
  We evaluate the performance of DeepThin inference across three Haswell- and Broadwell-based platforms with varying cache sizes.
  Speedups range from 2X to 14X, depending on the compression ratio and platform cache sizes.
\end{abstract}



\maketitle

\input{Intro}
\input{Arch}

\input{Library}
\input{Result}

\input{Conclusion}
\bibliography{cites.bib}



\end{document}

%% file: Intro.tex
\section{Introduction and Motivation}
In recent years, machine learning algorithms have been increasingly used in consumer-facing
products, such as speech recognition in personal assistants.
These algorithms rely on large weight
matrices which encode the relationships between different nodes in a network.
Ideally, these algorithms would run directly on the client devices
such as Amazon Echo~\cite{echo} and Google Home~\cite{googleHome}, which utilize them.
Unfortunately, because such devices are usually portable, low-power devices, running
such expensive algorithms is infeasible due to the significant storage, performance, and
energy limitations involved.

To work around this problem, many developers have resorted to executing the inference
models on high-performance cloud servers and streaming the inputs and outputs of the model
between client and server. This solution, however, introduces many
issues, including high operational costs, use of large
amounts of data transfer on metered (mobile) networks, user privacy concerns, and
increased latency.

Recent research has investigated methods of compressing models to sizes which can be
efficiently executed directly on the client device. Such compression approaches must
reduce the model space requirement without significantly impacting prediction accuracy,
runtime performance, or engineering time.
Our work builds on existing research in the area of low rank factorization.
We develop a new compression method and library, DeepThin, that:

\begin{enumerate}
\item Solves the fundamental symmetry issue with extremely low-rank matrix factorization of machine
  learning model parameters, using an auxiliary intermediate matrix and an efficient re-layout operation.
\item Is integrated with the popular and commonly used TensorFlow framework, enabling users 
    to seamlessly compress models at different granularities. We also implemented previous work
  compression techniques in our library to compare accuracy loss across compression methods. 
\item Consistently results in better accuracy at the same size as the previous compression methods.
\item With our customized C++ TensorFlow operations built on top of MKL~\cite{mkl}, empirically demonstrates
  inference performance speed-ups from $\mathrm{2X}$ to $\mathrm{14X}$ over uncompressed models.
\end{enumerate}

\section{Related Work}
One of the most successful approaches for reducing the total number of free
parameters in a network is iterative pruning \cite{iterprune}. A large network is initially
trained, then the subset of connections with magnitude closest to 0 are
removed. This train-prune cycle is repeated until the network reaches a desired size and
accuracy. Han et al. show iteratively pruned models can compress up to $13X$ with no loss of
accuracy \cite{iterprune}. Iterative pruning, however, is less than ideal in practice.

First, there is no way of storing the pruned network that fully realizes the
reduction in free parameters. Common storage methods for sparse matrices, including
compressed-sparse-row and -column formats~\cite{CSR}, more than double the actual stored size
of the network when compared to the number of free parameters. CSR, for example,
stores sparse matrices as a combination of three vectors - two of which hold one element
for each non-zero element in the matrix. Thus, a network pruned to $\mathrm{\frac{1}{N}}$ of
its original number of free parameters will be more than $\mathrm{\frac{2}{N}}$ of its original size once
it is actually stored in CSR form.

Furthermore, pruning works by taking advantage of the fact that the
vast majority of weights in a network are unimportant to the final output. After
removing all of those weights, however, further pruning forces the network to remove
increasingly important connections. This causes highly-pruned models to lose accuracy
rapidly and, empirically, limits the effectiveness of pruning when targeting
compressed sizes significantly smaller than $\mathrm{\frac{1}{50}}$ of the original model.

Finally, pruning networks introduces many hyperparameters (such as the intervals at which to
prune and the exact saliency algorithm to use \cite{betterpruning}), optimization of which can
have positive effects on compressed model accuracy but require significant hand-tuning during
the training process. Often this effort is not worth the additional accuracy gains, leaving many
pruned models with remaining inefficiency.

Alternatively, success has been found by reducing the bit width of and quantizing the network
weights \cite{deepcompression}. However, lowering bit widths imposes a large lower bound
on the compressed size (after which weights are represented by single bits and cannot be
quantized further).

Better encoding methods, particularly Huffman~\cite{huffman} and run-length coding, can build on
quantized weights to achieve significantly smaller file sizes \cite{deepcompression}.
Unfortunately, such techniques rely on the actual distribution of trained weights in a network,
making it impossible to accurately predict model sizes until training is
complete. This limitation is often unacceptable in practice, when one wishes to train a model
for specific size requirements. Additionally, such methods require significant overhead during
the layer computation (as they must decode each individual weight), further degrading model
performance.

HashedNetworks are similar to weight quantization and clustering, except the assignment
of weight to cluster is determined according to a hash function (so only the cluster means, and
not the weight-cluster mapping, need to be stored) \cite{hashednet}. Essentially, each 
element of the weight matrix is chosen at computation time from a vector of possible values
according to the hash of its row and column indices.

Such HashedNets require the computation of a hash function for each weight in the network,
adding significant computational complexity to a model. Furthermore, they rely on random memory
access patterns which are particularly difficult to optimize for. Finally, because weights are
shared randomly in a layer, it is difficult for HashedNetworks to learn logical local patterns
in data (which are particularly present in speech, image, and other continuous data).

Recently, Ha et al. introduced HyperNetworks, which use a small ``internal" network
to generate the weights for the main ``outer" network \cite{hypernet}. Ha et al. focus
on the power of such HyperNetworks to change their weights over time, but the
method also has significant compression potential. However, they hand-designed different,
complex internal networks for each class of model, making it more difficult to apply their
research to new model architectures without significant engineering
overhead.

Finally, Denil et al. and others have shown that weight matrices may be factored
into smaller matrices at training time to aid in distributed model training
\cite{predictingparams,ibmrankfact,convfactor}. Conceptually, weight factorization can be thought
of as a special case of the HyperNetwork idea, where the ``internal" network
is represented by a simple multiplication of the two matrix factors. Unfortunately, Denil et al.
were unable to effectively train both of the factored weight matrices at once, and instead
relied on a pretraining scheme to fix the values of one of the factored matrices before
training. These pretrained matrix values, while not needing to be further trained, are not
efficiently distributable, significantly limiting the technique's effectiveness beyond
improving training efficiency on distributed systems. Additionally, they did not provide a
means of further training the initialized matrix values, which should theoretically result in
improved accuracy.

This paper builds on the weight factorization design, improving on and empirically testing
it in many important ways to get our ``DeepThin" compression method. In addition, we provide
empirical and theoretical best practices for using DeepThin.

%% file: Arch.tex
\section{DeepThin Compression Model}
Standard deep neural networks consist of conceptual ``layers'' chained one after
another, through which input data passes sequentially until finally reaching a
desired output. Each layer computes a matrix multiplication between the outputs of
the previous layer and the current layer's weight matrix. After computing the
matrix multiplication, bias terms are added and a non-linear activation function
is applied to the output.

For data with some time dependency, recurrent neural networks (RNNs)
can be used. Although there are different types of RNNs, they all involve a model
containing a number (usually three or four) of compute steps similar to the one
described above. Such models can be more parameter-efficient
than regular DNNs, but they still require prohibitively large weight matrices to achieve
useful levels of accuracy and thus also benefit greatly from compression methods.

For visual data, convolutional neural networks (CNNs) sweep learned
filter banks (weights) over the input data to extract common features. Each sweep
step is computationally similar to the layer operation described above.
That said, because the size of input and output buffers in current state-of-the-art
convolutional networks represent an unusually large percent of actual network memory
requirements (due to large number of input/output channels), in this work we mainly focus on RNNs and feed forward DNNs.
Nevertheless, there is no fundamental limitation to apply DeepThin compression method to CNNs.

\begin{figure}[t!]
  \centering
  \includegraphics[width=0.82\linewidth]{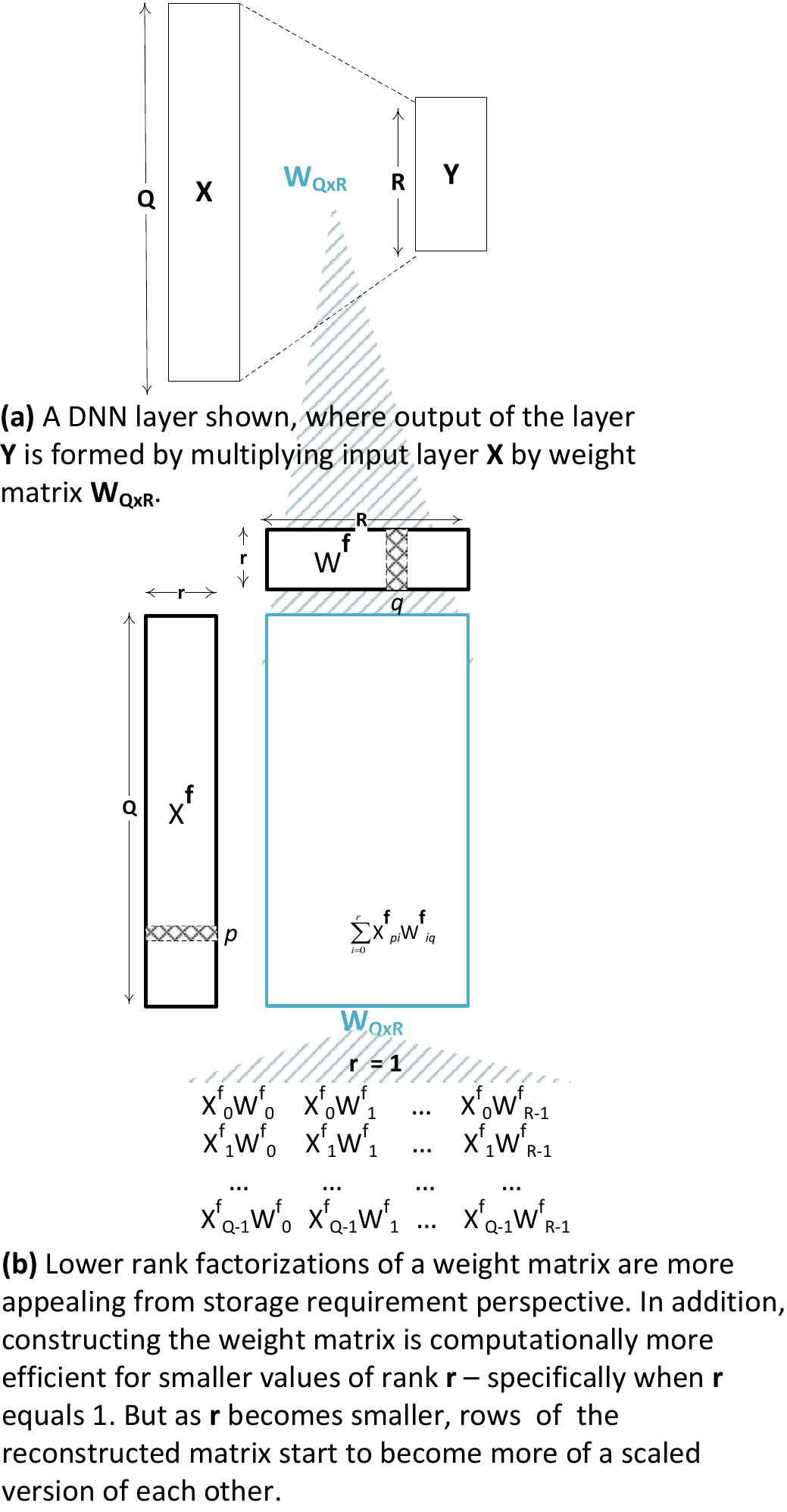}
  \caption{Lower rank factorization of a weight matrix: as r becomes smaller, rows and columns of the
    reconstructed matrix start to become more of a scaled version of each other.}
  \label{fig:arch:rank}
\end{figure}

Throughout this work we apply the compression methods to each layer's weight matrix
independently. A single layer with a non-linear activation function $\mathrm{a}$,
weights $\mathrm{W}$, and biases $\mathrm{B}$ is defined as:
\begin{equation}
  \label{eq:layer}
  \mathrm{Y = a(X.W + B)}
\end{equation}

, where $\mathrm{W}$ and $\mathrm{B}$ are learnable parameters that must be stored
within the network. As the size of $\mathrm{B}$ is often negligible compared to $\mathrm{W}$,
we only consider here compression of the $\mathrm{W}$ parameter (although we do compress the
biases in our evaluations).

The DeepThin architecture proposed in this paper can compress (with some loss of accuracy)
any model that relies on storing large weight matrices such as $\mathrm{W}$ in Equation~\ref{eq:layer}.

\section{DeepThin Architecture}
\label{sec:deepThin}

In this section we introduce our compression approach and show how it evolved from
a low-rank approximation method.

\begin{figure}[t!]
  \centering
  \includegraphics[width=0.8\linewidth]{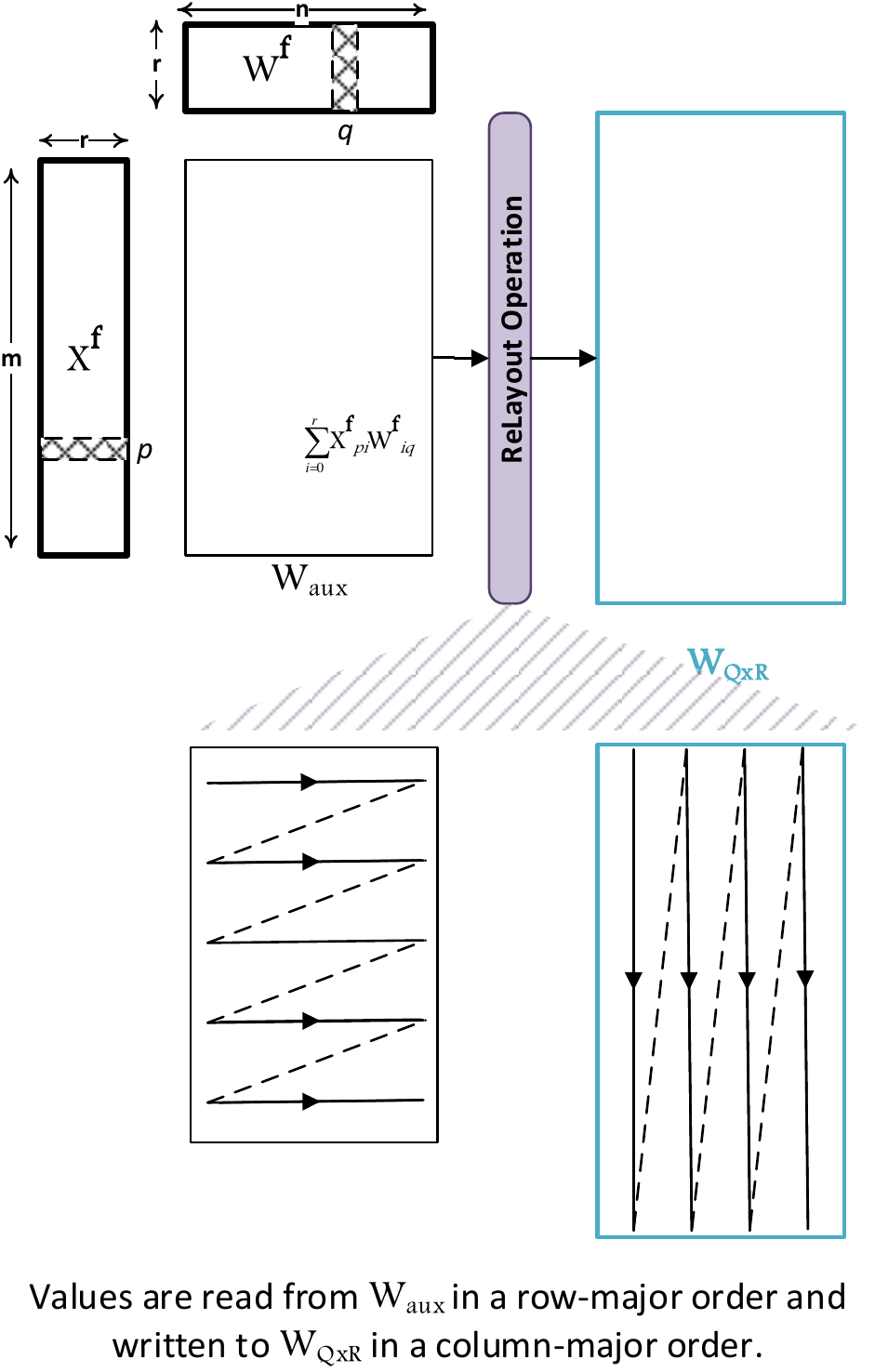}
  \caption{Breaking the artificial structural constraint created by factorization. This transform
    has two learnable parameters: low rank factors $\mathrm{X^f}$ and $W^f$.}
  \label{fig:arch:reshape}
\end{figure}

Rank factorization compression algorithms work by replacing the weight parameter $\mathrm{W}$
in Equation~\ref{eq:layer} with a function of a smaller set of learnable parameters \cite{predictingparams,hypernet,ibmrankfact}.
For example, in rank approximation as shown in Figure~\ref{fig:arch:rank}, a $\mathrm{Q\times R}$
weight matrix $\mathrm{W}$ is represented as the product of two lower rank matrices:

\begin{equation}
  \mathrm{W_{Q\times R} \approx X^f.W^f}
\end{equation}

where $\mathrm{X^f}$ is a $\mathrm{Q\times r}$ matrix and $\mathrm{W^f}$ is an $\mathrm{r\times R}$
matrix. During training, the error signal is back-propagated to the low rank factors $\mathrm{X^f}$
and $\mathrm{W^f}$ to update their elements the same way a regular weight matrix is trained. The
learned factors are then used to reconstruct the weight matrix at each layer during the
forward training or inference passes. Lower rank factorizations of a weight matrix -- specifically when
r equals 1 -- are more appealing from both a storage and computational efficiency standpoint. But
as $\mathrm{r}$ becomes smaller, rows/columns of the reconstructed weight matrix begin to resemble
each other. In the bottom part of Figure~\ref{fig:arch:rank} we show that, for r=1, every set (row)
of weights generated by the low rank approximation is a semi-scaled copy of the weight vector $\mathrm{W^f}$, and all columns are a scaled copy of the vector $\mathrm{X^f}$.
Unfortunately, we have found that this artificial resemblance considerably impacts the learning performance and capacity
of a network by requiring all output nodes in a layer to recognize very similar input patterns (introducing extreme redundancy).


To reduce the negative impact that such artificial constraints have on network learning capacity,
DeepThin first applies rank approximation to an auxiliary matrix $\mathrm{W_{aux}}$ of size
$\mathrm{m\times n}$.

\begin{equation}
  \label{eq:comp}
  \mathrm{W_{aux} = X^f.W^f}
\end{equation}

where now $\mathrm{X^f}$ is a $\mathrm{m\times r}$ matrix and $\mathrm{W^f}$
is an $\mathrm{r\times n}$ matrix.

Next, elements of $\mathrm{W_{aux}}$ are redistributed into $\mathrm{W_{Q\times R}}$
such that the artificial symmetry is broken. The reshaping process adds non-linearity to the approximation
function the same way an activation function adds a nonlinear decision boundary to the network's output layer.
On first glance, an ideal nonlinear transfer function would randomly scatter elements of matrix $\mathrm{W_{aux}}$
into matrix $\mathrm{W_{Q\times R}}$. We performed tests with that scattering as a limit study, and as expected
found that it was extremely slow (due primarily to the random memory accesses involved). Furthermore, such a scattering
scheme requires a matrix of indices to relate the original position to the re-laid-out
position, which undermines the compressibility. 

A more cost-effective alternative would be to distribute rows of $\mathrm{W_{aux}}$
along columns of $\mathrm{W_{Q\times R}}$ as shown in Figure~\ref{fig:arch:reshape}.
Note that this differs from a simple transpose in that one column of $\mathrm{W_{Q\times R}}$
may be composed of multiple (or only part of a) row of $\mathrm{W_{aux}}$.

This reshaping is especially appealing when computing a matrix multiplication because it
constructs a column (or part of a column) of $\mathrm{W_{Q\times R}}$ at a time, in an array of
consecutive memory location, while multiplying them by different rows of input matrix
$\mathrm{X}$ in Equation~\ref{eq:layer}. Thus, it makes full use of the generated elements
before discarding them.

Additionally, and to our surprise, we have found that reshaping with this method often achieves
better accuracy than with the random scatter method. See Figures~\ref{fig:tfvox:1} and~\ref{fig:dspch:1}
for a comparison between DeepThin and the random scattering (DeepThin Shuffled).
We hypothesize that this is due to the random scattering not being able to take advantage of local patterns in the data and weight matrices.
Thus, we find that there may be some trade-off between allowing the network to take advantage of such local
patterns and getting ``stuck" by forcing the usage of such patterns as rigidly as in the simple
factorization approach.

Nevertheless, the above reshape function may still result in repetition patterns in matrix $\mathrm{W_{Q\times R}}$
in form of blocks (of columns) scaled slightly differently. To solve this problem, it is important in practice
to choose the number of columns in matrix $\mathrm{W^f}$ to be prime with respect to the number of rows in matrix
$\mathrm{W_{Q\times R}}$. In other words, we prefer to have:

\begin{equation}
  \label{eq:prime}
  \mathrm{LCM(n, Q) = n \times Q}
\end{equation}

where LCM is the least common multiple of the two numbers. The LCM value determines
the repetition frequency of similar, scaled blocks in matrix $\mathrm{W_{Q\times R}}$.
Here $\mathrm{Q}$, the original weight matrix width, is fixed. So, ideally we must
choose the largest value for $\mathrm{n}$ that is prime with respect to $\mathrm{Q}$.
But parameter $\mathrm{n}$ must also satisfy other constraints within the DeepThin compression
framework. First, matrix $\mathrm{W_{aux}}$ needs to have at least as many elements
as matrix $\mathrm{W_{Q\times R}}$. Therefore:

\begin{equation}
  \label{eq:limit}
  \mathrm{m\times n \ge Q\times R}
\end{equation}


In addition, DeepThin parameters need to abide by the specified compression rate. Because DeepThin
compresses weights on a per-matrix basis, we can calculate the compressed size of any individual
compressed matrix in the network. Therefore, for a compression ratio of $\mathrm{\alpha}$
and a weight matrix of shape $Q\times R$ we have:

\begin{equation}
  \label{eq:ratio}
  \mathrm{\alpha = \frac{r\times (m+n)}{Q\times R}}
\end{equation}

where the denominator totals the size of the weight matrix $\mathrm{W_{Q\times R}}$, and
the enumerator sums up the size of low rank factors of matrix $\mathrm{W_{aux}}$. Note that we
do not include the size of the bias vector in this calculation. In practice, we compress and
compute the bias vector's compression rate separately if required by the user.
Again, note that Equation~\ref{eq:ratio} ignores other, much smaller parameters of the network that are not compressed
(such as batch-normalization~\cite{batchnorm} parameters), thus the actual network size will have some overhead.
If the exact network architecture is known, one may add those other parameters to Equation~\ref{eq:ratio}
to get an exact ratio for the entire network or to further compress the other parameters to reach
a desired overall compress rate.
Our library automatically compensates for these parameters, but for ease
of explanation we maintain the simple form of Equation~\ref{eq:ratio}.

Now, if we replace $\mathrm{m}$ in Equation~\ref{eq:ratio} with its lower bound value derived from
Inequality~\ref{eq:limit}, for each rank factorization value ($\mathrm{r=1, r=2, \cdot}$), we have
a single variable quadratic inequality as shown below:

\begin{equation}
  \label{eq:alpha}
  \mathrm{\alpha \ge \frac{r\times (\frac{Q\times R}{n}+n)}{Q\times R}}
\end{equation}

for which valid ranges of $\mathrm{n}$ can be easily determined if any exist. Within valid ranges,
we then pick a
minimum value of $\mathrm{n}$ that satisfies conditions expressed by Equations~\ref{eq:limit}
and~\ref{eq:prime}. Since this fine tuning only happens once during initialization of the network
the performance overhead is negligible.

Non-existence of a valid range for $\mathrm{n}$ in Inequality~\ref{eq:alpha} indicates that
the spatial overhead of DeepThin surpasses any compression
benefit it provides for the specific matrix being compressed. This creates an effective
lower-bound on the compression rate $\mathrm{\alpha}$ supported by DeepThin. This lower-bound
is dependent on the size of the original network weight matrix $\mathrm{W_{Q\times R}}$ but
is often less than $\mathrm{\frac{1}{1000}}$ of the matrix size.

Additionally, different matrices in a network may have different shapes and thus different
lower bounds. This makes it often possible to over-compress larger matrices in order to
``make up for" matrices that have hit this lower bound, thus achieving a desired overall
compressed network size (see Section~\ref{sec:lib:target} for our library's approach to this).

\subsection{Similarities to a Single-Layer Neural Network}
Conceptually, the DeepThin architecture can be thought of as an inner,
single-layer neural network that generates the weights for the larger outer
network layer (akin to a HyperNetwork). This single-layer ``internal"
network does not include a
bias or activation function, though it benefits from nonlinearity
added via the re-layout transformation. Although biases serve an important
role in standard network layers, allowing the network to effectively shift
the activation function, our ``internal" networks do not need such biases for multiple reasons.
First, values in $\mathrm{W^f}$ and $\mathrm{X^f}$ are
distributed around a mean of $\mathrm{0}$. So any biases added to the output
of the DeepThin transform would determine the expected mean of the generated
weights $\mathrm{W_{Q\times R}}$. Because original network weights, $\mathrm{W_{Q\times R}}$,
are almost always centered about $0$, removing the biases from the internal
network does not affect accuracy.
Additionally, because we do not have activation
functions on the ``internal" network, biases do not provide the ``function-shifting"
benefit. Activation functions, on the other hand, have squashing effects that may
limit the range of values generated for the reconstructed matrix
$\mathrm{W_{Q\times R}}$ -- a new set of artificial constraints that may impact the
learning capacity of the original network.

Single-layer internal networks have many benefits, including requiring very
little engineering and better compute efficiency at runtime. Nevertheless, a
natural question is whether a deeper model could help better reconstruct matrix
$\mathrm{W_{Q\times R}}$, and improve the compressed network's accuracy.

Ideally, one should be able to extend the DeepThin architecture
such that lower rank factor matrices -- $\mathrm{X^{f}}$ and $\mathrm{W^{f}}$ in Figure~\ref{fig:arch:reshape}
are themselves recursively lower-rank approximated and reshaped within another inner DeepThin layer.
The premise here is that only the inner-most layer's thinner rank factor matrices are learned and
stored with the hope that they can reconstruct much larger weight matrices on the fly for the
outer-most DeepThin layer(s), resulting in lower loss of accuracy.

We performed experiments with such extensions of DeepThin, and were able to show an
experimental accuracy improvement of up to only about $\mathrm{1\%}$. This soon hits the
point of diminishing returns, as deeper internal networks become significantly more
difficult to optimize for (particularly taking into account the size of intermediate
buffers that would need to be preserved).
Therefore, in the rest of the paper, we focus on the single layer DeepThin architecture.
In Section~\ref{sec:res:acc} we show that the single-layer DeepThin performs very well,
almost always beating other compression methods.


%% file: Library.tex
\section{DeepThin Library Implementation}



We implement DeepThin and previously proposed compression methods as library modules integrated with the open-source TensorFlow~\cite{tensorflow}
framework. The TensorFlow framework first builds a graph of learnable variables
and operations on those variables in Python. Next, the graph is evaluated
on the compute device using platform-specific libraries. TensorFlow
is capable of automatically differentiating complex graphs built with a standard
set of operations included in TensorFlow. Furthermore, optimization algorithms included
with TensorFlow can use automatically calculated gradients to iteratively optimize
all the learnable variables on a graph, with the details of gradient computation and parameter
updates abstracted from the user or other libraries.

\subsection{Training}
For training, instead of calling the standard TensorFlow library to declare learnable
variables for each of the network's weight matrices, the user calls a function in our library
which mimics the TensorFlow API. Instead of declaring a single learnable variable node on
the computation graph, our library creates a separate sub-graph for each variable,
which may have its own learnable parameters (e.g., $\mathrm{X^f}$ and $\mathrm{W^f}$, for DeepThin-compressed matrices) and
operations (re-layout operation defined in
Section~\ref{sec:deepThin}). The output of that sub-graph can then be used as a drop-in
replacement for a regular learnable matrix in TensorFlow. Additionally, because we are able
to express our operations as a combination of existing TensorFlow operations, TensorFlow is
able to automatically derive the gradient computations and correctly backpropagate the error
signal to the learnable parameters of the sub-graph ($\mathrm{X^f}$ and $\mathrm{W^f}$ for DeepThin).

For DeepThin-compressed models, the sub-graph consists of two learnable parameters, $\mathrm{X^f}$ and
$\mathrm{W^f}$, which undergo a matrix-multiplication, transposition, slice, and reshape in order to
implement our re-layout operation.

The rank factorization sub-graph is simpler, where two factors $\mathrm{X^f}$ and $\mathrm{W^f}$ are multiplied
and the result is used in the rest of the graph. Note that the first dimension of $\mathrm{X^f}$ and the
last dimension of $\mathrm{W^f}$ must match the respective dimensions of the requested matrix shape,
a requirement that is not applied to DeepThin-compressed models, where the re-layout operation
reshapes and slices the product matrix $\mathrm{W_{aux}}$ to the desired size.

HashedNet sub-graphs contain a learnable set of ``hash bins", then a fixed mapping from bins to locations
in the uncompressed matrix $\mathrm{W}$ (which is determined by pre-computing the hash as described in the
HashedNet paper\cite{hashednet} and source code). Finally, a gather operation is performed
and the resulting $\mathrm{W}$ matrix is used in the model graph.

Pruned matrices are created by a sub-graph that consists of a learnable matrix and a ``mask"
composed of binary 0 and 1 values. The two are element-wise multiplied, and the resultant
masked matrix is returned. The library user must place a call to our library's ``step" function
within their training loop, to indicate that another training iteration has passed.
At each training iteration, the library determines whether or not to prune the network at this
step -- determined by a user-configured pruning schedule.
The mask starts off filled with 1s, and each time the network is pruned some percentage of the elements in the mask are
switched to 0 based on the magnitude of their associated weight in the learnable matrix. By the end of the
last pruning step the mask has exactly enough 1s such that when
stored in CSR format the respective non-zero parameters would take up the targeted amount of storage
space.

To the programmer, this entire process is transparent and consists only of
changing a few function calls to use our library (such as replacing \texttt{tf.get\_tensor}
with \texttt{dt.get\_tensor}). Furthermore, training works exactly as before, and the
optimization functions included with TensorFlow will correctly backpropagate to the learnable
parameters within each sub-graph.

Because same-size networks are extremely
model-dependent, we do not support that compression method in our library. Instead, the user
tells the library to return an uncompressed model, and he or she manually modifies the relevant
hyper-parameters until it is satisfactorily compressed. We added this
configuration to facilitate comparison with hand-optimized compressed models.

\subsection{Initialization of Internal Parameters}
Our library initializes the learnable parameters of each sub-graph such that
the overall distrubution of the represented variable (weight matrix elements) is preserved.
For example, for DeepThin-compressed networks, the learnable parameters $\mathrm{X^f}$ and $\mathrm{W^f}$
must be initialized such that entries in the reconstructed matrix $\mathrm{W_{Q\times R}}$ hold
initial distributions similar to those of the uncompressed model. To produce reconstructed weights with
initial variance $\mathrm{\sigma^2}$, our initialization scheme is as follows:

\begin{itemize}
\item Initialize entries of rank factor $\mathrm{X^f}$ with the same variance, $\sigma^2$, as the original weights should have
\item Initialize all entries of rank factor  $\mathrm{W^f}$ to a variance of $\mathrm{\frac{1}{r}}$ to maintain that $\sigma^2$ variance through the reconstruction process
\end{itemize}
The above initialization scheme works since, in neural networks, weights are randomly
initialized by a zero-mean distribution. Weight $\mathrm{w_{j,k}}$ in the reconstructed matrix
is computed by the following sum:

\begin{equation}
  \mathrm{\sum_{i=0}^{r-1} X^f_{j,i}W^f_{i,k}}
\end{equation}

Since $\mathrm{X^f_{j,i}}$ and $\mathrm{W^f_{i,k}}$ are independent random variables with
mean of zero, variance of their product equals the product of their variance,
$\mathrm{\frac{\sigma^2}{r}}$. This summed over $\mathrm{r}$ terms results in the target
variance of $\mathrm{\sigma^2}$ for $\mathrm{w_{j,k}}$.
Similar calculations are performed to derive initial mean and standard deviations
for non-0-mean initialized variables.

\subsection{Targeting Compressed Sizes}
\label{sec:lib:target}
For all compression methods supported by our library, we set a maximum
compressed size ratio and had the library automatically calculate the 
compression parameters that would best achieve that size target.

Note that, because DeepThin imposes a lower-bound on compression rate determined
per-matrix based on the matrix's shape, targeting an overall network compression rate
is not as simple as just compressing all learnable parameters to that ratio (as smaller
parameters may hit the lower bound before reaching that ratio).
To work around this, we utilize a multi-pass approach, where the user first defines
the entire computation graph as well as the desired overall compression rate. In this first
step, we leave the shapes of compressing sub-graph parameters undeclared, as we do not yet know to what
extent we can or wish to compress each parameter. Next, our library attempts to set the sizes
of the sub-graph parameters such that all matrices are compressed to the desired
compression ratio, keeping track of when matrices reached their lower-bound, and the delta
between the desired size and the lower bound. Finally, we attempt to distribute the
``overhead'' caused by matrices hitting their lower-bounds (and the small number of non-compressed model parameters) among other matrices which have not
yet reached the lower-bound.
This approach allows us to very accurately target compressed sizes to the desired maximum size.

Again, the same-size neural network method
is model-dependent and was thus not included in our library. Instead, we hand-tuned
the number of hidden units to roughly match the size of the other compression methods.
Note that for networks more complex than simple feed-forward networks,
deciding exactly how to manually reduce network size is a highly involved process,
as deep combinations of convolutional, residual, and other layer types
often do not have a single parameter which directly controls the network
size.


\subsubsection{Inference}
\label{sec:lib:inf}
\begin{figure*}[t!]
  \centering
  \includegraphics[width=1.0\linewidth]{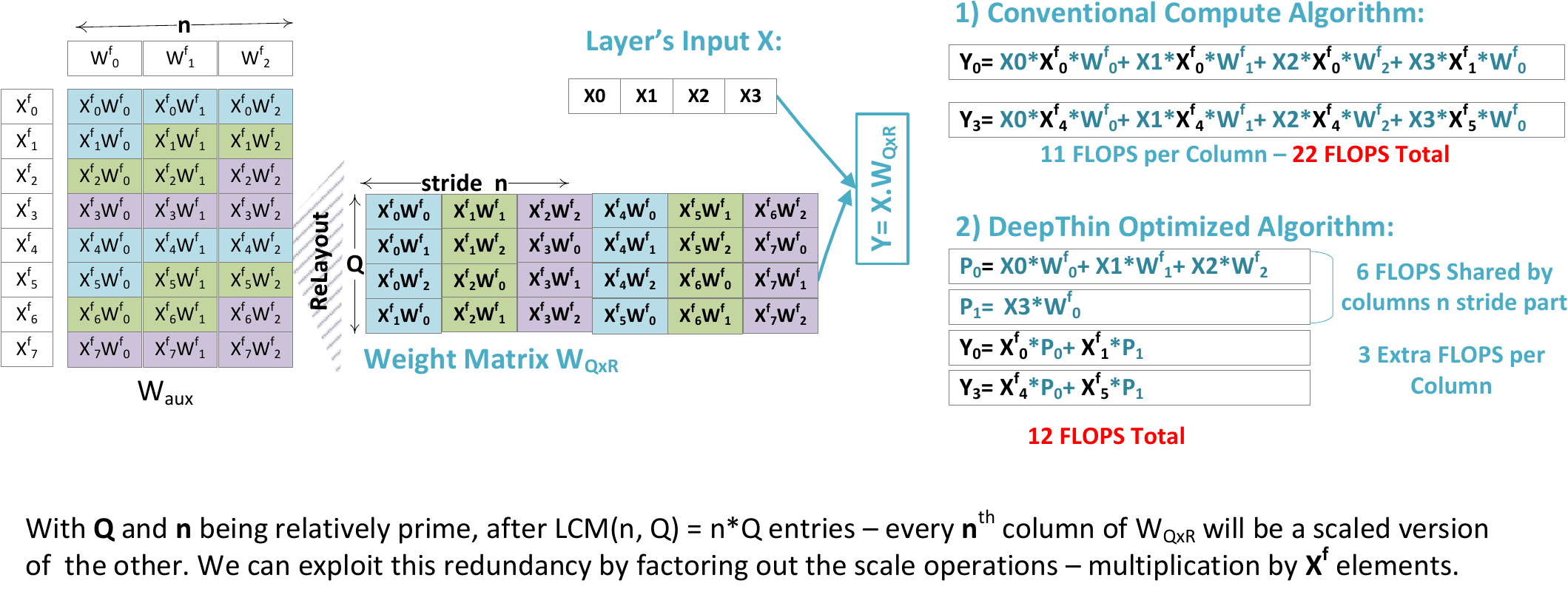}
  \caption{Exploiting redundancy generated by weight compression to optimize computation of $\mathrm{Y=X.W_{QxR}}$.}
  \label{fig:arch:reuse}
\end{figure*}

We tested multiple compression methods and find experimentally in Section~\ref{sec:res:acc}
that our proposed DeepThin compression method results in better accuracy than any competing method.
Motivated
by these findings, we decided to investigate the performance of DeepThin-compressed networks
further by writing a custom kernel for inference (rather than using existing TensorFlow operations)
that computes a DeepThin-compressed matrix multiplication
directly, without explicitly decompressing the entire weight matrix at once. We hope that such
a fused operation will significantly lower runtime memory, compute, and energy usage of the inference phase
by fitting the entire model in the compute device's cache and taking advantage of particular features of the DeepThin
compression method to reduce total computation.

Such considerations are extremely important, especially as devices that are storage-constrained
enough to need a compressed model are also very likely to be
memory-, compute-, or power-constrained as well. Computing such matrix multiplications more
efficiently during the inference phase
can directly lead to faster experiences and better battery life, two major
considerations on mobile devices.

We wrote our operation directly in C++, using MKL~\cite{mkl} for computations and interfacing
with the TensorFlow C++ APIs to define our kernel as a valid operation which can be added to a
TensorFlow graph as described above. Our operation takes three parameters: the low-rank factors
$\mathrm{X^f}$ and $\mathrm{W^f}$, as well as the layer input to be multiplied. During the training phase, we tested a range of
values for $\mathrm{r}$, and found that $\mathrm{r = 1}$ works either as well or better than other possible
values with regard to accuracy, and as such all future discussion will assume $\mathrm{r = 1}$.

As demonstrated in Figure~\ref{fig:arch:reuse}, we can think of the decompressed weight matrix
$\mathrm{W}$ as having copies of $\mathrm{W^f}$
tiled in a column-major fashion
throughout the matrix. Each tiled copy of $\mathrm{W^f}$ is scaled by the corresponding scalar in
$\mathrm{X^f}$. The matrix multiplication is computed such that each cell in the output matrix
consists of the sum of the dot products between each of the $\mathrm{X^f}$-scaled copies of $\mathrm{W^f}$ in a single
column and the respective slices of the input $\mathrm{X}$.
We consider two major optimizations for the above DeepThin-compressed matrix multiplication.

First, instead of scaling each copy of $\mathrm{W^f}$ by an element of $\mathrm{X^f}$ and \textit{then} computing
the dot product against the input slice (requiring $\mathrm{2\times n}$ total multiply-adds),
one can \textit{first} compute the dot product between the relevant portions of $\mathrm{W^f}$ and the input slice, then scale the
resulting scalar value(s) by the scalar elements of $\mathrm{X^f}$ (requiring $\mathrm{n + 1}$ multiply-adds).

Second, following the above approach, we recognize that at certain points in the
computation, shown in Figure~\ref{fig:arch:reuse}, we will end up computing the dot
product between the same elements of $\mathrm{W^f}$ and the same input slice (although that partial product will be later
scaled by different elements from $\mathrm{X^f}$). This redundancy can be exploited, such that after the
first time a particular $\mathrm{W^f}$ dot product is computed for a particular input slice that dot
product is stored, and the kernel scales that dot product by all the $\mathrm{X^f}$ values using it and
sums those partial products for all of the cells that make use of that same dot product. This can
significantly cut down on the most expensive part of the matrix multiplication, the dot
product, by reusing these repetitive computations. Note that the number of times a particular
dot product can be reused decreases as the LCM between the $\mathrm{Q}$ and $\mathrm{n}$ dimensions defined in
Section~\ref{sec:deepThin} increases. Additionally, larger $\mathrm{R}$ dimensions leads to more overall
computation, but also more possibilities for reuse.




%% file: Result.tex
\input{Result-Accuracy}
\input{Result-Perf}

%% file: Result-Accuracy.tex
\section{Accuracy Results}
\label{sec:res:acc}
We compare the accuracy of models compressed with DeepThin
for the same compression rate against previous deep compression techniques, which include:
\begin{itemize}
    \item HashedNet: A small set of possible weight values are distributed into a larger
    weight matrix according to a hashing function~\cite{hashednet}.
    \item Pruning: Matrices are iteratively pruned and the resultant sparse matrices
    stored with CSR format until the overall size is achieved~\cite{iterprune}.
    \item Same-Size Networks: Manually lowering the number of hidden units in a network until the desired
      size is reached. 
    \item Rank Factorization: Factoring weight matrices into smaller, lower-rank factors~\cite{predictingparams,ibmrankfact}.
\end{itemize}
For DeepThin, all networks were compressed with $\mathrm{r = 1}$ and $\mathrm{n}$ calculated as discussed
in Section~\ref{sec:deepThin}.
We also compare against DeepThin-Shuffled -- a DeepThin-compressed network, where the weight values are randomly
distributed throughout the final weight matrix (as opposed to our re-layout operation).

We compared the accuracy of all compression methods on two state-of-the-art speech recognition models.
All accuracy plots have had models with a very similar compression rate grouped together horizontally
around their mean compression rate in order to increase readability (as each compression method
may have its own small differences in the final compressed size). Additionally, rank factorization
tends to have a higher lower-bound size than the other methods (due to the requirement that the
factor matrices retain dimensions $Q$ and $R$), and as such have fewer possible
configurations.

Although preliminary results show DeepThin accuracy is
very competitive on ResNet and other CNN models, We do
not report results for CNN models because the unusually
large number of input/output channels and activations in modern CNNs
would prevent inference models from fitting on the target devices even
if the weights were completely removed (although there is
no fundamental limitation of our method).




\subsection{TFKaldi/Voxforge}
First, we evaluated all compression methods on an acoustic model modified from the commonly used open-source speech recognition toolkit
TFKaldi~\cite{tfkaldi} and trained on the free Voxforge dataset~\cite{voxforge}. The model
is a standard feed-forward network consisting of $6$ hidden layers of $2048$ units each.
The inputs to the model are $440$ pre-processed audio features, each representing a frame
of recorded speech. The outputs of the model are approximately $6928$ ``phones;" sub-syllable
units of speech that are later compiled into words using a grammar-aware beam search
decoder (we do not attempt to compress the decoder and do not include it in our size
calculations, as it is not part of the deep network). We used the standard Kaldi~\cite{kaldi}
decoder to decode the phones and calculate word error rate (WER). As is standard in the
TFKaldi toolkit, we report the lowest WER result out of a series of test batches.

The TFKaldi model contains $36,080,400$ parameters and has an application (speech recognition)
that is often used on low-power devices, making it a realistic target for compression.

\begin{figure}[t!]
    \includegraphics[width=1.0\linewidth]{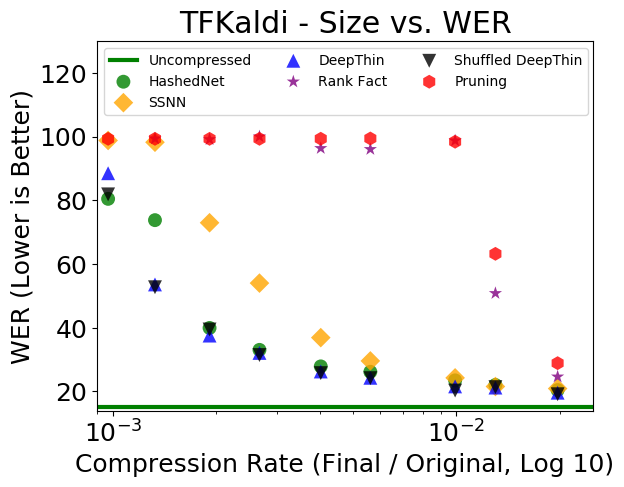}
    \caption[Size vs. WER - TFKaldi model]
    {
        Comparing the test word error rate (WER) of different compression methods at a range
        of compressed sizes for the TFKaldi model. Note that DeepThin models result in lower
        WERs at all compression rates tested except for the very smallest,
        $\mathrm{\frac{1}{1000}}$, configuration, for which HashedNet achieves a lower WER.
        However, HashedNetworks have prohibitively expensive computational requirements and
        memory access patterns. Pruning and rank factorization rapidly
        degrade in performance as the model becomes smaller. Same-size networks perform well,
        but even with the non-scalable manual effort fail to match our low error rates.
    }
    \label{fig:tfvox:1}
\end{figure}

Figure~\ref{fig:tfvox:1} compares the test Word Error Rate (WER) at the end of five training
epochs for each compression method compared. As expected, DeepThin networks show better results
than competing methods at practically all tested compression rates. The average relative
improvement in WER compared to each other method is summarized in Table~\ref{table:acc:1}.

Looking closer at Figure~\ref{fig:tfvox:1}, we first compare DeepThin to a same-size network,
which represents a hand-tuned reduction in model parameters by reducing the size of each hidden
layer in the network. We find that DeepThin consistently outperforms such same-size networks,
clearly showing that our method makes more efficient use of its parameters even at such small
sizes.

Next, we find that pruned models perform particularly poorly, losing nearly all ability to
learn at compressed sizes less than approximately $\mathrm{\frac{1}{100}}$ of the uncompressed
network. This can be attributed to the requirement that vast majorities
of the model parameters be set to $0$ in order to maintain the compression rate under
CSR. This gives very little freedom to the network to learn anything at all. The inefficiencies
of CSR only exacerbate the problem as the sizes get extremely small, requiring even more
parameters to be pruned away than the compression rate would suggest.

The HashedNet model at the very smallest configuration has a slightly lower WER than
the corresponding DeepThin networks, however such HashedNet models require significantly
more computation and randomized memory access patterns which are very difficult to optimize
for. Additionally, at all other configurations the DeepThin models outperform HashedNet, by
up to $2-3\%$ absolute WER.

Rank factorization performs very poorly as the size decreases, which is mostly due
to the increase in symmetry effects discussed in Section~\ref{sec:deepThin} as the rank
of the factors decreases to reach a lower compressed size. Solving this symmetry issue
with the auxiliary matrix and re-layout operation as described in Section~\ref{sec:deepThin}
is one of our main contributions, and the improvements from rank factorization to the DeepThin
models visible in Figure~\ref{fig:tfvox:1} highlight its success.

Finally, we see that our method performs very similarly to a ``shuffled DeepThin" approach,
where $\mathrm{W_{aux}}$ is randomly distributed into $\mathrm{W}$ (thus completely avoiding any symmetries in
the final weight matrix). This shows that our re-layout operation, while being significantly
more efficient to compute than a random shuffle, is still able to effectively deal with the 
symmetry issues caused by such factorization methods.

\begin{table}[b!]
  \begin{tabular}{ |c|c|c|c|c| }
   
 \hline
 & Rank & Prune & SSNN & Hash \\
\hline
TFKaldi & $60.08\%$ & $56.96\%$ & $23.40\%$ & $6.12\%$ \\
\hline
DeepSpeech & $28.09\%$ & $27.21\%$ & $20.45\%$ & $12.16\%$ \\
\hline
\end{tabular}

\caption[Average Relative Improvement From Using DeepThin]
{
    Average relative improvement of DeepThin compared to four other compression methods.
    TFKaldi numbers are in relative WER decrease, while DeepSpeech numbers are in relative
    test error reduction. Here we see that different compression methods tend to do better on
    different datasets, however DeepThin consistently beats all other compression methods tested.
}
\label{table:acc:1}
\end{table}

\begin{figure}[t!]
    \includegraphics[width=1.05\linewidth]{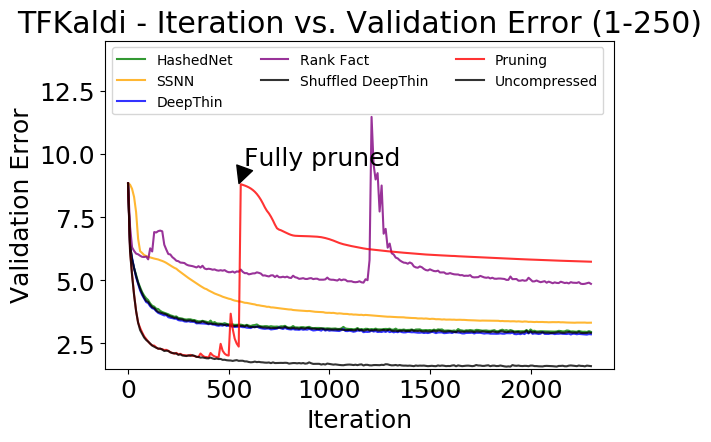}
    \caption[Iteration vs. Validation Loss - TFKaldi model]
    {
        Tracking the validation loss for a single compression rate ($\frac{1}{250}$)
        on the TFKaldi model, compared across the different methods tested.
        Note that the pruned model is not completely compressed until iteration 550,
        at which point it trains very poorly. Additionally, we see instability
        in the training of rank-factored matrices which is removed in the DeepThin
        curve through the use of the re-layout operation. HashedNets have very similar
        accuracy results to the DeepThin method, but rely on random memory access patterns
        and complex hash computations that significantly impede model runtime performance.
    }
    \label{fig:tfvox:2}
\end{figure}

Figure~\ref{fig:tfvox:2} shows the change in validation loss during training for a single
compressed size (about $\mathrm{\frac{1}{250}}$ of the uncompressed model). Note that
the pruned network (in red) begins training as an uncompressed model, then prunes weights
incrementally every 50 iterations until the 550th iteration.
As expected, we find that pruning has very little effect on accuracy for
the first few pruning iterations, during which time it prunes away almost exclusively
unimportant weights that would be essentially $0$ even in a dense network. After
approximately step 400, however, the prune steps are visible as large jumps in the
validation loss, as the network is forced to begin removing weights that are vital
to accurately predicting the output classes.

This visually demonstrates both how well pruning works for fairly large compressed
network sizes (the initial smooth, decreasing portion of the graph where the weights
pruned are unimportant) and the negative effects of pruning past this point (when the
validation loss begins to increase significantly after each pruning step).

Rank-factorization (simply factoring each $Q\times R$ matrix
into a $\mathrm{Q\times r}$ and $\mathrm{r\times R}$ matrix)
is particularly unstable. Multiple times throughout the training process it diverges, and
eventually ends up as the second-worst performing model. As the target compressed size gets
smaller, we have found that this instability becomes an even larger issue, preventing it from
ever reaching a satisfactory convergence better than the initial random parameters. This issue is
caused primarily by the symmetries described in Section~\ref{sec:deepThin}, which motivated and is solved
by the use of an auxiliary matrix and re-layout operation in DeepThin.

We see as well that, although the same-size network approach reaches a reasonable
validation loss by the end of training, it takes significantly longer to converge than 
DeepThin or HashedNet-compressed networks, which is not ideal when training resources are in
high demand.

Finally, Figure~\ref{fig:tfvox:2} shows that HashedNets have extremely similar training
curves to the DeepThin models, but our DeepThin models ultimately converge to a lower validation
loss and WER. Additionally, HashedNets require both the computation of a complex hash and a
random memory access for each individual weight in a weight matrix, which can have extreme
effects on runtime performance. In contrast, we show in Section~\ref{sec:results:perf} that
DeepThin has no such issues, and can be highly optimized to run significantly faster than
uncompressed models, especially with the product-reuse technique demonstrated in 
Figure~\ref{fig:arch:reuse}.

\subsection{DeepSpeech}
We also tested DeepThin and other discussed compression techniques on a state-of-the-art speech recognition system called
DeepSpeech~\cite{deepspeech}. This model, recently open sourced by Baidu, combines a one-dimensional convolutional
layer, multiple GRU recurrent layers~\cite{GRU}, and a feed-forward layer
to produce a character prediction from a raw frequency spectrum. This system can be
trained end-to-end to directly predict text from speech at the character level without
a decoder.

We had a number of problems with the open-source model, particularly with
its tendency to overfit the limited training data available. To overcome
this, and to integrate it with our library, we ported it to TensorFlow,
added learning rate decay, and removed batch normalization~\cite{batchnorm} on the
recurrent layers.

The open-source DeepSpeech model contains neither a language model back-end nor any word error
rate evaluation code, which prohibits us from reporting model WER.
Instead, we report the test loss
using the evaluation scripts included with the model.

\begin{figure}[t!]
    \includegraphics[width=1.0\linewidth]{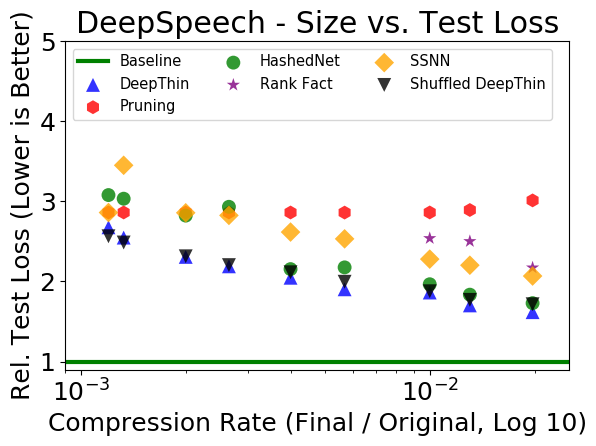}
    \caption[Size vs. Error - DeepSpeech model]
    {
        Comparing the test loss of different compression methods at a range of compressed
        model sizes for the DeepSpeech speech recognition model. Note here that DeepThin models
        significantly outperform all other compression methods, while same-size networks perform
        well for larger models but fail when the compressed size is made smaller. Pruning, rank
        factorization, and HashedNets all quickly devolve into ineffective models, only ever
        predicting the few most common letters (or complete silence) instead of learning
        useful models.
    }
    \label{fig:dspch:1}
\end{figure}

Figure~\ref{fig:dspch:1} compares the best test loss after twenty training epochs
for each compression method compared on the DeepSpeech model. Here, DeepThin networks
outperform all other compression methods at all tested configurations. The average relative
improvement in test loss compared to each other method is summarized in
Table~\ref{table:acc:1}.

Furthermore, Figure~\ref{fig:dspch:1} shows that pruning performs particularly poorly
for this model. We believe that because the model contains a number of smaller matrices,
pruned models are forced to prune many layers almost completely, and this issue compounds
itself as the input goes through the series of heterogeneous layers, each layer receiving
significantly less information.

Same-size models perform competitively when the model size is relatively large, but lose
almost all learning capacity as the sizes decreases.

HashedNets are more consistent, and do not lose learning ability as much as same-size or pruned
networks, however even at the larger size networks tested they perform worse than DeepThin
networks. Coupled with the serious performance and optimization issues encountered with
HashedNets, they are clearly not a good choice for this model.

The rank-factorization models, meanwhile, show mediocre results and hit their lower-bound 
at approximately $\mathrm{\frac{1}{100}}$ the original model size - making them very
difficult to compare to other methods and impossible to use for the majority of compression
ratios targeted.

Furthermore, as with TFKaldi, we find that our proposed method performs slightly better than a shuffled DeepThin approach. Again, this demonstrates the effectiveness of our re-layout operation in
removing symmetry issues while remaining highly efficient to compute. We also expect the
non-shuffled DeepThin to do particularly well here, as it can still take advantage of local
patterns in the input data that are particularly strong for the raw frequency spectrum inputs of
DeepSpeech compared to the pre-processed features in the TFKaldi model.

\subsection{LCM Effects}
\label{subsec:LCM}
Next, we wanted to compare the effects of the LCM between $\mathrm{n}$ and $\mathrm{Q}$
as discussed in Section~\ref{sec:deepThin}.

\begin{figure}[b!]
   \includegraphics[width=1\linewidth]{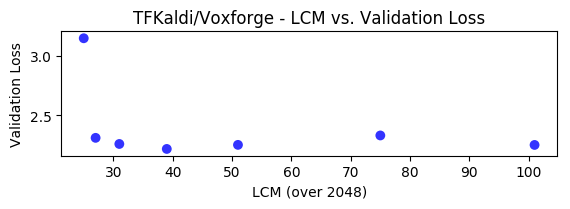}
   \caption[LCM vs. Validation Loss]
     {
       Compares the accuracy of a model at multiple LCM values. Note that LCM values are 
       reported with respect to the hidden and output layers that have an input dimension of
       $2048$ units, and the actual LCMs have been divided by $2048$ (the largest common factor)
       to improve readability. Here we see that not optimizing LCM values can cause large accuracy
       issues, but beyond an initial optimization the trend is less clear. It is clear, however, that
       maximizing the LCM is a reasonable approach to minimize the validation loss without testing multiple
       complete configurations and training rounds.
     }
   \label{fig:lcm}
\end{figure}

We did this by comparing the same compression rate (approximately $\mathrm{\frac{1}{100}}$)
on TFKaldi using multiple different $\mathrm{n}$ values, to achieve different LCMs. The result
is shown in Figure~\ref{fig:lcm}.

Here we can clearly see the positive effect of choosing values for $\mathrm{n}$ that increase
the LCM, and find a $\mathrm{28.56\%}$ relative decrease in validation loss by increasing the LCM (divided by $2048$)
from $25$ to $101$.

However, the trend is not extremely strong - as long as the LCM is not actively
\textit{minimized}, it seems that there is some variation in final validation loss given a
certain LCM. This is partly because the figure is incomplete - in such a simple experiment, we
can only consider the LCM with respect to one set of layers at a time, in this case the majority of
the layers which have an input of $2048$ units. At the same time, the first layer has $440$
input units, which has a different factorization and causes different LCMs with the same values
of $\mathrm{n}$. This means that many of the points have higher or lower LCMs with respect to
the input layer than they do with the other layers shown in the figure, causing extra symmetry
issues that affect the final validation loss which are not accounted for in the figure. This
highlights the importance of choosing a different $\mathrm{n}$ for each layer with a different
input dimension $\mathrm{Q}$ in order to maximize the LCM on a per-layer basis, which our
library is able to do automatically.

Additionally, we believe that the association between higher LCM and lower validation loss is
mostly a rule of thumb - as long as the LCM is reasonably large, further increases impact
model capacity to a lesser extent.

%% file: Result-Perf.tex
\section{Performance Results}
\label{sec:results:perf}
We evaluated the performance of DeepThin inference across three
Haswell- and Broadwell-based platforms with varying cache sizes.  Both the baseline, uncompressed implementations and DeepThin leverage MKL.
The first platform is a Broadwell machine (CONF1) with 4 cores/8
threads at 3.30 GHz, 1 memory channel, and 6 MB of L3 cache.
Next, we have a Broadwell-E machine (CONF2) with 10 cores/20 threads at 3.00 GHz, 8
memory channels, and 25 MB of L3 cache.
Finally, the third platform is a Haswell machine (CONF3) with 36 cores/72 threads running at 2.30 GHz, 16 memory channels, and an L3 cache
of 45 MB.  
All machines had 32 KB of L1 and 256 KB of L2 cache, and all tests were run on the CPU only.

We were particularly interested in how much faster our DeepThin models are when compared
to the original, uncompressed model. To test this, we ran 1,000 one-batch iterations through
the uncompressed TFKaldi DNN,
and 100 one-batch iterations through the uncompressed DeepSpeech network (DeepSpeech is a 
considerably slower network, so we ran fewer batches). After recording the time taken, we
ran the same tests on DeepThin-compressed versions of each model using the custom TensorFlow
operation described in Section~\ref{sec:lib:inf} and optimized to compute directly on the low-rank factors
$\mathrm{X^f}$ and $\mathrm{W^f}$. We report the speedup (X) for each configuration
in Table~\ref{table:perf:1}.


\begin{table}
  \small 
  \begin{tabular}{ |p{.85cm}|p{.85cm}|p{.85cm}|p{.85cm}|p{.85cm}|p{.85cm}|p{.85cm}| }
 \hline
 & \multicolumn{3}{|l|}{TFKaldi} & \multicolumn{3}{|l|}{DeepSpeech} \\
 \hline
  \textasciitilde Rel. Size & \tiny CONF1 & \tiny CONF2 & \tiny CONF3
                            & \tiny CONF1 & \tiny CONF2 & \tiny CONF3 \\
\hline
0.0195& 3.80X& 2.51X& 2.46X
& 7.66X&  2.24X& 2.08X \\

0.0129& 5.64X& 3.16X& 3.51X
& 10.66X& 3.09X& 2.77X \\

0.0099& 6.47X& 3.88X& 4.09X
& 12.12X& 3.50X& 3.09X \\

0.0057& 8.12X& 4.21X& 4.84X
& 13.69X& 3.89X& 3.67X \\

0.0040& 8.53X& 5.69X& 5.32X
& 13.72X& 3.86X& 3.70X \\

0.0027& 8.22X& 5.22X& 5.42X
& 13.04X& 3.65X& 3.46X \\

0.0020& 7.43X& 4.56X& 4.58X
& 11.68X& 3.36X& 3.22X \\

0.0014& 4.44X& 4.12X& 2.44X
& 8.72X&  2.57X& 2.42X \\
\hline
\end{tabular}
\caption[DeepThin Speed Improvements vs. Uncompressed Model]
{
    Comparing the execution speed of DeepThin models across both TFKaldi and DeepSpeech at
    a range of different compressed sizes and machines. Compressed sizes have small variations
    between machines, but are accurate within $0.0001$. All results are presented in ``X faster"
    than the speed of the uncompressed, baseline model. We find that the largest gains come from
    platforms with smaller cache sizes, but that using DeepThin consistently reduces execution
    time across all configurations tested, making it ideal for environments where latency and
    power usage are important.
}
\label{table:perf:1}
\end{table}

Table~\ref{table:perf:1} clearly demonstrates the impressive performance gains realizable
from using DeepThin, sometimes reaching speeds of up to $14X$ faster than the uncompressed model.
This effect is primarily visible on the machine equipped with smaller caches that more closely represents a real-world
lower-end client situation, though we also
find large performance benefits on even the most capable, cluster machine.

Most of these performance gains come from a smaller overall memory footprint that often results
in the entire model being stored in the CPU cache(s), significantly improving performance
over multiple runs, as it no longer has to wait to receive model parameters from the slower
main memory.

However, a smaller memory footprint alone is not enough - models like HashedNet, for example,
have similarly small memory usage but require the computation of an expensive hash function
and a random, non-contiguous memory access to compute the matrix multiplication. DeepThin, by
contrast, requires no such hashing function and uses completely contiguous memory accesses.
Furthermore, as discussed in Section~\ref{sec:lib:inf}, DeepThin models can be
further optimized for inference performance by re-using partial products, which can cut the amount
of computation by over $2\times$ in some models.

Note that the trend in the table is that smaller models run faster, up to about $0.0027$ of
the original model size. 
However, once the model is small enough to fit in a sufficiently-fast cache, 
the most important factor in the models speed quickly becomes how often the dot products
can be reused, as shown in Figure~\ref{fig:arch:reuse}. As the models
get smaller, $\mathrm{n}$ must increase (see Equation~\ref{eq:alpha} - $\mathrm{n}$ is usually
much smaller than $\mathrm{Q\times R}$, making the first term in the numerator - which shrinks as n 
increases - the most important in determining compression rate), which
means there are fewer total partial products and a larger LCM, allowing for less total reuse
in the computation and thus more overall computation.

%% file: Conclusion.tex
\section{Conclusions}
Compressing machine learning models to run on storage-, compute-, and power-constrained
devices is quickly becoming a major area of research and industry interest. Existing compression
methods struggle to compress models below $1-2\%$ of their original sizes and/or add significantly
to the computational complexity of the models. We propose DeepThin, a compression method improving
significantly on low-rank matrix factorization. We also present a compression library that integrates
with TensorFlow to quickly and easily compress machine learning models with multiple compression methods.
We evaluate DeepThin and find that it achieves up to $60\%$ better results
than competing methods. Finally, we devise a custom kernel for DeepThin-compressed networks
and demonstrate inference speed increases of up to $14X$ over uncompressed networks
using a variety of performance optimizations.

%% file: paper.bbl

\begin{thebibliography}{20}


\ifx \showCODEN    \undefined \def \showCODEN     #1{\unskip}     \fi
\ifx \showDOI      \undefined \def \showDOI       #1{#1}\fi
\ifx \showISBNx    \undefined \def \showISBNx     #1{\unskip}     \fi
\ifx \showISBNxiii \undefined \def \showISBNxiii  #1{\unskip}     \fi
\ifx \showISSN     \undefined \def \showISSN      #1{\unskip}     \fi
\ifx \showLCCN     \undefined \def \showLCCN      #1{\unskip}     \fi
\ifx \shownote     \undefined \def \shownote      #1{#1}          \fi
\ifx \showarticletitle \undefined \def \showarticletitle #1{#1}   \fi
\ifx \showURL      \undefined \def \showURL       {\relax}        \fi
\providecommand\bibfield[2]{#2}
\providecommand\bibinfo[2]{#2}
\providecommand\natexlab[1]{#1}
\providecommand\showeprint[2][]{arXiv:#2}

\bibitem[\protect\citeauthoryear{??}{vox}{2017}]%
        {voxforge}
 \bibinfo{year}{August 2017}\natexlab{}.
\newblock \bibinfo{title}{{VoxForge} English Corpus}.
\newblock   (\bibinfo{year}{August 2017}).
\newblock
\showURL{%
\url{http://www.repository.voxforge1.org/downloads/SpeechCorpus/Trunk/Audio/Main/}}


\bibitem[\protect\citeauthoryear{Abadi, Agarwal, Barham, Brevdo, Chen, Citro,
  Corrado, Davis, Dean, Devin, Ghemawat, Goodfellow, Harp, Irving, Isard, Jia,
  Jozefowicz, Kaiser, Kudlur, Levenberg, Man\'{e}, Monga, Moore, Murray, Olah,
  Schuster, Shlens, Steiner, Sutskever, Talwar, Tucker, Vanhoucke, Vasudevan,
  Vi\'{e}gas, Vinyals, Warden, Wattenberg, Wicke, Yu, and Zheng}{Abadi
  et~al\mbox{.}}{2015}]%
        {tensorflow}
\bibfield{author}{\bibinfo{person}{Mart\'{\i}n Abadi}, \bibinfo{person}{Ashish
  Agarwal}, \bibinfo{person}{Paul Barham}, \bibinfo{person}{Eugene Brevdo},
  \bibinfo{person}{Zhifeng Chen}, \bibinfo{person}{Craig Citro},
  \bibinfo{person}{Greg~S. Corrado}, \bibinfo{person}{Andy Davis},
  \bibinfo{person}{Jeffrey Dean}, \bibinfo{person}{Matthieu Devin},
  \bibinfo{person}{Sanjay Ghemawat}, \bibinfo{person}{Ian Goodfellow},
  \bibinfo{person}{Andrew Harp}, \bibinfo{person}{Geoffrey Irving},
  \bibinfo{person}{Michael Isard}, \bibinfo{person}{Yangqing Jia},
  \bibinfo{person}{Rafal Jozefowicz}, \bibinfo{person}{Lukasz Kaiser},
  \bibinfo{person}{Manjunath Kudlur}, \bibinfo{person}{Josh Levenberg},
  \bibinfo{person}{Dan Man\'{e}}, \bibinfo{person}{Rajat Monga},
  \bibinfo{person}{Sherry Moore}, \bibinfo{person}{Derek Murray},
  \bibinfo{person}{Chris Olah}, \bibinfo{person}{Mike Schuster},
  \bibinfo{person}{Jonathon Shlens}, \bibinfo{person}{Benoit Steiner},
  \bibinfo{person}{Ilya Sutskever}, \bibinfo{person}{Kunal Talwar},
  \bibinfo{person}{Paul Tucker}, \bibinfo{person}{Vincent Vanhoucke},
  \bibinfo{person}{Vijay Vasudevan}, \bibinfo{person}{Fernanda Vi\'{e}gas},
  \bibinfo{person}{Oriol Vinyals}, \bibinfo{person}{Pete Warden},
  \bibinfo{person}{Martin Wattenberg}, \bibinfo{person}{Martin Wicke},
  \bibinfo{person}{Yuan Yu}, {and} \bibinfo{person}{Xiaoqiang Zheng}.}
  \bibinfo{year}{2015}\natexlab{}.
\newblock \bibinfo{title}{{TensorFlow}: Large-Scale Machine Learning on
  Heterogeneous Systems}.
\newblock   (\bibinfo{year}{2015}).
\newblock
\showURL{%
\url{https://www.tensorflow.org/}}
\newblock
\shownote{Software available from tensorflow.org.}


\bibitem[\protect\citeauthoryear{Amodei, Ananthanarayanan, Anubhai, Bai,
  Battenberg, Case, Casper, Catanzaro, Cheng, Chen, Chen, Chen, Chen,
  Chrzanowski, Coates, Diamos, Ding, Du, Elsen, Engel, Fang, Fan, Fougner, Gao,
  Gong, Hannun, Han, Johannes, Jiang, Ju, Jun, LeGresley, Lin, Liu, Liu, Li,
  Li, Ma, Narang, Ng, Ozair, Peng, Prenger, Qian, Quan, Raiman, Rao, Satheesh,
  Seetapun, Sengupta, Srinet, Sriram, Tang, Tang, Wang, Wang, Wang, Wang, Wang,
  Wang, Wu, Wei, Xiao, Xie, Xie, Yogatama, Yuan, Zhan, and Zhu}{Amodei
  et~al\mbox{.}}{2016}]%
        {deepspeech}
\bibfield{author}{\bibinfo{person}{Dario Amodei}, \bibinfo{person}{Sundaram
  Ananthanarayanan}, \bibinfo{person}{Rishita Anubhai},
  \bibinfo{person}{Jingliang Bai}, \bibinfo{person}{Eric Battenberg},
  \bibinfo{person}{Carl Case}, \bibinfo{person}{Jared Casper},
  \bibinfo{person}{Bryan Catanzaro}, \bibinfo{person}{Qiang Cheng},
  \bibinfo{person}{Guoliang Chen}, \bibinfo{person}{Jie Chen},
  \bibinfo{person}{Jingdong Chen}, \bibinfo{person}{Zhijie Chen},
  \bibinfo{person}{Mike Chrzanowski}, \bibinfo{person}{Adam Coates},
  \bibinfo{person}{Greg Diamos}, \bibinfo{person}{Ke Ding},
  \bibinfo{person}{Niandong Du}, \bibinfo{person}{Erich Elsen},
  \bibinfo{person}{Jesse Engel}, \bibinfo{person}{Weiwei Fang},
  \bibinfo{person}{Linxi Fan}, \bibinfo{person}{Christopher Fougner},
  \bibinfo{person}{Liang Gao}, \bibinfo{person}{Caixia Gong},
  \bibinfo{person}{Awni Hannun}, \bibinfo{person}{Tony Han},
  \bibinfo{person}{Lappi Johannes}, \bibinfo{person}{Bing Jiang},
  \bibinfo{person}{Cai Ju}, \bibinfo{person}{Billy Jun},
  \bibinfo{person}{Patrick LeGresley}, \bibinfo{person}{Libby Lin},
  \bibinfo{person}{Junjie Liu}, \bibinfo{person}{Yang Liu},
  \bibinfo{person}{Weigao Li}, \bibinfo{person}{Xiangang Li},
  \bibinfo{person}{Dongpeng Ma}, \bibinfo{person}{Sharan Narang},
  \bibinfo{person}{Andrew Ng}, \bibinfo{person}{Sherjil Ozair},
  \bibinfo{person}{Yiping Peng}, \bibinfo{person}{Ryan Prenger},
  \bibinfo{person}{Sheng Qian}, \bibinfo{person}{Zongfeng Quan},
  \bibinfo{person}{Jonathan Raiman}, \bibinfo{person}{Vinay Rao},
  \bibinfo{person}{Sanjeev Satheesh}, \bibinfo{person}{David Seetapun},
  \bibinfo{person}{Shubho Sengupta}, \bibinfo{person}{Kavya Srinet},
  \bibinfo{person}{Anuroop Sriram}, \bibinfo{person}{Haiyuan Tang},
  \bibinfo{person}{Liliang Tang}, \bibinfo{person}{Chong Wang},
  \bibinfo{person}{Jidong Wang}, \bibinfo{person}{Kaifu Wang},
  \bibinfo{person}{Yi Wang}, \bibinfo{person}{Zhijian Wang},
  \bibinfo{person}{Zhiqian Wang}, \bibinfo{person}{Shuang Wu},
  \bibinfo{person}{Likai Wei}, \bibinfo{person}{Bo Xiao}, \bibinfo{person}{Wen
  Xie}, \bibinfo{person}{Yan Xie}, \bibinfo{person}{Dani Yogatama},
  \bibinfo{person}{Bin Yuan}, \bibinfo{person}{Jun Zhan}, {and}
  \bibinfo{person}{Zhenyao Zhu}.} \bibinfo{year}{2016}\natexlab{}.
\newblock \showarticletitle{Deep Speech 2 : End-to-End Speech Recognition in
  English and Mandarin}. In \bibinfo{booktitle}{{\em Proceedings of The 33rd
  International Conference on Machine Learning}} {\em
  (\bibinfo{series}{Proceedings of Machine Learning Research})},
  \bibfield{editor}{\bibinfo{person}{Maria~Florina Balcan} {and}
  \bibinfo{person}{Kilian~Q. Weinberger}} (Eds.), Vol.~\bibinfo{volume}{48}.
  \bibinfo{publisher}{PMLR}, \bibinfo{address}{New York, New York, USA},
  \bibinfo{pages}{173--182}.
\newblock
\showURL{%
\url{http://proceedings.mlr.press/v48/amodei16.html}}


\bibitem[\protect\citeauthoryear{Chen, Wilson, Tyree, Weinberger, and
  Chen}{Chen et~al\mbox{.}}{2015}]%
        {hashednet}
\bibfield{author}{\bibinfo{person}{Wenlin Chen}, \bibinfo{person}{James~T.
  Wilson}, \bibinfo{person}{Stephen Tyree}, \bibinfo{person}{Kilian~Q.
  Weinberger}, {and} \bibinfo{person}{Yixin Chen}.}
  \bibinfo{year}{2015}\natexlab{}.
\newblock \showarticletitle{Compressing Neural Networks with the Hashing
  Trick}. In \bibinfo{booktitle}{{\em Proceedings of the 32Nd International
  Conference on International Conference on Machine Learning - Volume 37}} {\em
  (\bibinfo{series}{ICML'15})}. \bibinfo{publisher}{JMLR.org},
  \bibinfo{pages}{2285--2294}.
\newblock
\showURL{%
\url{http://dl.acm.org/citation.cfm?id=3045118.3045361}}


\bibitem[\protect\citeauthoryear{Cho, Merrienboer, Bahdanau, and Bengio}{Cho
  et~al\mbox{.}}{2014}]%
        {GRU}
\bibfield{author}{\bibinfo{person}{Kyunghyun Cho}, \bibinfo{person}{Bart~Van
  Merrienboer}, \bibinfo{person}{Dzmitry Bahdanau}, {and}
  \bibinfo{person}{Yoshua Bengio}.} \bibinfo{year}{2014}\natexlab{}.
\newblock \showarticletitle{On the Properties of Neural Machine Translation:
  Encoder–Decoder Approaches}.
\newblock \bibinfo{journal}{{\em Proceedings of SSST-8, Eighth Workshop on
  Syntax, Semantics and Structure in Statistical Translation\/}}
  (\bibinfo{year}{2014}).
\newblock
\showDOI{%
\url{https://doi.org/10.3115/v1/w14-4012}}


\bibitem[\protect\citeauthoryear{Denil, Shakibi, Dinh, Ranzato, and
  de~Freitas}{Denil et~al\mbox{.}}{2013}]%
        {predictingparams}
\bibfield{author}{\bibinfo{person}{Misha Denil}, \bibinfo{person}{Babak
  Shakibi}, \bibinfo{person}{Laurent Dinh},
  \bibinfo{person}{Marc\textquotesingle~Aurelio Ranzato}, {and}
  \bibinfo{person}{Nando de Freitas}.} \bibinfo{year}{2013}\natexlab{}.
\newblock \showarticletitle{Predicting Parameters in Deep Learning}.
\newblock In \bibinfo{booktitle}{{\em Advances in Neural Information Processing
  Systems 26}}, \bibfield{editor}{\bibinfo{person}{C.~J.~C. Burges},
  \bibinfo{person}{L.~Bottou}, \bibinfo{person}{M.~Welling},
  \bibinfo{person}{Z.~Ghahramani}, {and} \bibinfo{person}{K.~Q. Weinberger}}
  (Eds.). \bibinfo{publisher}{Curran Associates, Inc.},
  \bibinfo{pages}{2148--2156}.
\newblock
\showURL{%
\url{http://papers.nips.cc/paper/5025-predicting-parameters-in-deep-learning.pdf}}


\bibitem[\protect\citeauthoryear{Ha, Dai, and Le}{Ha et~al\mbox{.}}{2016}]%
        {hypernet}
\bibfield{author}{\bibinfo{person}{David Ha}, \bibinfo{person}{Andrew Dai},
  {and} \bibinfo{person}{Quoc Le}.} \bibinfo{year}{2016}\natexlab{}.
\newblock \showarticletitle{HyperNetworks}.
\newblock


\bibitem[\protect\citeauthoryear{Han, Mao, and Dally}{Han
  et~al\mbox{.}}{2016}]%
        {deepcompression}
\bibfield{author}{\bibinfo{person}{Song Han}, \bibinfo{person}{Huizi Mao},
  {and} \bibinfo{person}{William~J Dally}.} \bibinfo{year}{2016}\natexlab{}.
\newblock \showarticletitle{Deep Compression: Compressing Deep Neural Networks
  with Pruning, Trained Quantization and Huffman Coding}.
\newblock \bibinfo{journal}{{\em International Conference on Learning
  Representations (ICLR)\/}} (\bibinfo{year}{2016}).
\newblock


\bibitem[\protect\citeauthoryear{Han, Pool, Tran, and Dally}{Han
  et~al\mbox{.}}{2015}]%
        {iterprune}
\bibfield{author}{\bibinfo{person}{Song Han}, \bibinfo{person}{Jeff Pool},
  \bibinfo{person}{John Tran}, {and} \bibinfo{person}{William Dally}.}
  \bibinfo{year}{2015}\natexlab{}.
\newblock \showarticletitle{Learning both Weights and Connections for Efficient
  Neural Network}. In \bibinfo{booktitle}{{\em Advances in Neural Information
  Processing Systems (NIPS)}}. \bibinfo{pages}{1135--1143}.
\newblock


\bibitem[\protect\citeauthoryear{{Huffman}}{{Huffman}}{1952}]%
        {huffman}
\bibfield{author}{\bibinfo{person}{David~A. {Huffman}}.}
  \bibinfo{year}{1952}\natexlab{}.
\newblock \showarticletitle{A Method for the Construction of Minimum-Redundancy
  Codes}.
\newblock \bibinfo{journal}{{\em Proceedings of the IRE\/}}
  \bibinfo{volume}{40}, \bibinfo{number}{9} (\bibinfo{year}{1952}),
  \bibinfo{pages}{1098--1101}.
\newblock


\bibitem[\protect\citeauthoryear{Intel}{Intel}{2017}]%
        {mkl}
\bibfield{author}{\bibinfo{person}{Intel}.} \bibinfo{year}{2017}\natexlab{}.
\newblock \bibinfo{title}{Intel Math Kernel Library}.
\newblock   (\bibinfo{year}{2017}).
\newblock
\showURL{%
\url{https://software.intel.com/en-us/mkl}}


\bibitem[\protect\citeauthoryear{{Ioffe} and {Szegedy}}{{Ioffe} and
  {Szegedy}}{2015}]%
        {batchnorm}
\bibfield{author}{\bibinfo{person}{Sergey {Ioffe}} {and}
  \bibinfo{person}{Christian {Szegedy}}.} \bibinfo{year}{2015}\natexlab{}.
\newblock \showarticletitle{Batch Normalization: Accelerating Deep Network
  Training by Reducing Internal Covariate Shift}.
\newblock \bibinfo{journal}{{\em international conference on machine
  learning\/}} (\bibinfo{year}{2015}), \bibinfo{pages}{448--456}.
\newblock


\bibitem[\protect\citeauthoryear{{Jaderberg}, {Vedaldi}, and
  {Zisserman}}{{Jaderberg} et~al\mbox{.}}{2014}]%
        {convfactor}
\bibfield{author}{\bibinfo{person}{Max {Jaderberg}}, \bibinfo{person}{Andrea
  {Vedaldi}}, {and} \bibinfo{person}{Andrew {Zisserman}}.}
  \bibinfo{year}{2014}\natexlab{}.
\newblock \showarticletitle{Speeding up Convolutional Neural Networks with Low
  Rank Expansions.}. In \bibinfo{booktitle}{{\em British Machine Vision
  Conference 2014}}.
\newblock


\bibitem[\protect\citeauthoryear{Li, Sainath, Narayanan, Caroselli, Bacchiani,
  Misra, Shafran, Sak, Pundak, Chin, Sim, Weiss, Wilson, Variani, Kim, Siohan,
  Weintraub, McDermott, Rose, and Shannon}{Li et~al\mbox{.}}{2017}]%
        {googleHome}
\bibfield{author}{\bibinfo{person}{Bo Li}, \bibinfo{person}{Tara Sainath},
  \bibinfo{person}{Arun Narayanan}, \bibinfo{person}{Joe Caroselli},
  \bibinfo{person}{Michiel Bacchiani}, \bibinfo{person}{Ananya Misra},
  \bibinfo{person}{Izhak Shafran}, \bibinfo{person}{Hasim Sak},
  \bibinfo{person}{Golan Pundak}, \bibinfo{person}{Kean Chin},
  \bibinfo{person}{Khe~Chai Sim}, \bibinfo{person}{Ron~J. Weiss},
  \bibinfo{person}{Kevin Wilson}, \bibinfo{person}{Ehsan Variani},
  \bibinfo{person}{Chanwoo Kim}, \bibinfo{person}{Olivier Siohan},
  \bibinfo{person}{Mitchel Weintraub}, \bibinfo{person}{Erik McDermott},
  \bibinfo{person}{Rick Rose}, {and} \bibinfo{person}{Matt Shannon}.}
  \bibinfo{year}{2017}\natexlab{}.
\newblock \showarticletitle{Acoustic Modeling for Google Home}.
\newblock
\showURL{%
\url{http://www.cs.cmu.edu/~chanwook/MyPapers/b_li_interspeech_2017.pdf}}


\bibitem[\protect\citeauthoryear{{Povey}, {Ghoshal}, {Boulianne}, {Burget},
  {Glembek}, {Goel}, {Hannemann}, {Motlicek}, {Qian}, {Schwarz}, {Silovsky},
  {Stemmer}, and {Vesely}}{{Povey} et~al\mbox{.}}{2011}]%
        {kaldi}
\bibfield{author}{\bibinfo{person}{Daniel {Povey}}, \bibinfo{person}{Arnab
  {Ghoshal}}, \bibinfo{person}{Gilles {Boulianne}}, \bibinfo{person}{Lukas
  {Burget}}, \bibinfo{person}{Ondrej {Glembek}}, \bibinfo{person}{Nagendra
  {Goel}}, \bibinfo{person}{Mirko {Hannemann}}, \bibinfo{person}{Petr
  {Motlicek}}, \bibinfo{person}{Yanmin {Qian}}, \bibinfo{person}{Petr
  {Schwarz}}, \bibinfo{person}{Jan {Silovsky}}, \bibinfo{person}{Georg
  {Stemmer}}, {and} \bibinfo{person}{Karel {Vesely}}.}
  \bibinfo{year}{2011}\natexlab{}.
\newblock \showarticletitle{The Kaldi Speech Recognition Toolkit}. In
  \bibinfo{booktitle}{{\em IEEE 2011 Workshop on Automatic Speech Recognition
  and Understanding}}.
\newblock


\bibitem[\protect\citeauthoryear{{Reed}}{{Reed}}{1993}]%
        {betterpruning}
\bibfield{author}{\bibinfo{person}{Russell {Reed}}.}
  \bibinfo{year}{1993}\natexlab{}.
\newblock \showarticletitle{Pruning algorithms-a survey}.
\newblock \bibinfo{journal}{{\em IEEE Transactions on Neural Networks\/}}
  \bibinfo{volume}{4}, \bibinfo{number}{5} (\bibinfo{year}{1993}),
  \bibinfo{pages}{740--747}.
\newblock


\bibitem[\protect\citeauthoryear{Renkens}{Renkens}{2017}]%
        {tfkaldi}
\bibfield{author}{\bibinfo{person}{Vincent Renkens}.} \bibinfo{year}{May
  2017}\natexlab{}.
\newblock \bibinfo{title}{Train a tensorflow neural net with kaldi alignments
  as targets}.
\newblock   (\bibinfo{year}{May 2017}).
\newblock
\showURL{%
\url{https://github.com/vrenkens/tfkaldi}}


\bibitem[\protect\citeauthoryear{Sainath, Kingsbury, Sindhwani, Arisoy, and
  Ramabhadran}{Sainath et~al\mbox{.}}{2013}]%
        {ibmrankfact}
\bibfield{author}{\bibinfo{person}{Tara~N. Sainath}, \bibinfo{person}{Brian
  Kingsbury}, \bibinfo{person}{Vikas Sindhwani}, \bibinfo{person}{Ebru Arisoy},
  {and} \bibinfo{person}{Bhuvana Ramabhadran}.}
  \bibinfo{year}{2013}\natexlab{}.
\newblock \showarticletitle{Low-rank matrix factorization for Deep Neural
  Network training with high-dimensional output targets}.
\newblock \bibinfo{journal}{{\em 2013 IEEE International Conference on
  Acoustics, Speech and Signal Processing\/}} (\bibinfo{year}{2013}),
  \bibinfo{pages}{6655--6659}.
\newblock


\bibitem[\protect\citeauthoryear{Snay}{Snay}{1976}]%
        {CSR}
\bibfield{author}{\bibinfo{person}{R.A. Snay}.}
  \bibinfo{year}{1976}\natexlab{}.
\newblock \showarticletitle{Reducing the profile of sparse symmetric matrices}.
\newblock \bibinfo{journal}{{\em Bull. Geodesique\/}} (\bibinfo{year}{1976}).
\newblock


\bibitem[\protect\citeauthoryear{Weber and Alexa}{Weber and Alexa}{2016}]%
        {echo}
\bibfield{author}{\bibinfo{person}{Anthony Weber} {and} \bibinfo{person}{Amazon
  Alexa}.} \bibinfo{year}{2016}\natexlab{}.
\newblock \bibinfo{booktitle}{{\em Amazon Echo: The Best User Guide to Master
  Amazon Echo Fast}}.
\newblock \bibinfo{publisher}{CreateSpace Independent Publishing Platform},
  \bibinfo{address}{USA}.
\newblock
\showISBNx{1523893907, 9781523893904}


\end{thebibliography}
